\documentclass[twoside]{article}

\usepackage{aistats2019}
\usepackage{latexsym,amsmath,amssymb,amsthm,amsfonts,graphicx, subcaption, mathtools}

\usepackage{epsfig,bm,authblk,url}
\usepackage{caption}
\usepackage{floatrow}
\usepackage{diagbox}
\graphicspath{ {figures/} }
\usepackage{siunitx}
\usepackage{float} 
\usepackage{booktabs} 
\usepackage[margin=1cm]{caption} 
\usepackage{mathtools}
\usepackage[ruled]{algorithm2e}
\usepackage{algpseudocode}%


\usepackage{titling}
\newtheorem{assumption}{Assumption} 
\newtheorem{corollary}{Corollary} 
\newtheorem{proposition}{Proposition} 
\usepackage[title]{appendix} 
\usepackage{enumitem} 
\floatstyle{plain}
\newfloat{Algorithm}{thp}{lop}
\floatname{Algorithm}{Algorithm}

\newcommand{\Cov}{\mathrm{Cov}}

\newcommand{\EE}{\mathbb{E}}
\newcommand{\ELBO}{\mathcal{L}}
\newcommand{\Var}{\mathbb{V}}

\newcommand{\R}{\mathbb{R}}
\newcommand{\Q}{\mathbb{Q}}

\newcommand{\E}{\mathbb{E}}

\usepackage{scalerel,stackengine}
\stackMath
\newcommand\reallywidehat[1]{%
\savestack{\tmpbox}{\stretchto{%
  \scaleto{%
    \scalerel*[\widthof{\ensuremath{#1}}]{\kern-.6pt\bigwedge\kern-.6pt}%
    {\rule[-\textheight/2]{1ex}{\textheight}}
  }{\textheight}%
}{0.5ex}}%
\stackon[1pt]{#1}{\tmpbox}%
}

\usepackage{xr}

\usepackage[round]{natbib}

\begin{document}

%

%

\twocolumn[

\aistatstitle{Variance reduction properties of the reparameterization trick}

\aistatsauthor{Ming Xu$^\dagger$ \And Matias Quiroz$^\ddagger$ \And  Robert Kohn$^\ddagger$ \And  Scott A. Sisson$^\dagger$}
\aistatsaddress{$^\dagger$University of New South Wales \And $^\ddagger$UNSW Business School}] 



\begin{abstract}
 The reparameterization trick is widely used in variational inference as it yields more accurate estimates of the gradient of the variational objective than alternative approaches such as the score function method. Although there is overwhelming empirical evidence in the literature showing its success, there is relatively little research exploring why the reparameterization trick is so effective. We explore this under the idealized assumptions that the variational approximation is a mean-field Gaussian density and that the log of the joint density of the model parameters and the data is a quadratic function that depends on the variational mean. From this, we show that the marginal variances of the reparameterization gradient estimator are smaller than those of the score function gradient estimator. We apply the result of our idealized analysis to real-world examples.
\end{abstract}

\section{INTRODUCTION}\label{sec:Intro}

\paragraph{Background}

Variational inference (VI) \citep{jordan1999introduction, ormerod, BleiReview} provides a fast and approximate alternative to exact Monte Carlo methods when performing Bayesian inference on parameters in complex statistical models. The idea of VI is to approximate the posterior density with a family of tractable densities, indexed by variational parameters, where a member of that family is referred to as a variational approximation. VI then proceeds by finding a set of variational parameters such that the variational approximation is close to the true posterior density in some sense. In machine learning, VI has been used in generative models through variational autoencoders \citep{KingmaWelling}. In econometrics and statistics, complex regression density estimation \citep{villaninott}, state space models \citep{TanNott}, and high-dimensional time-varying parameter models \citep{QuirozNottKohn} are approximated using VI. Furthermore, VI has recently been extended to cases where the likelihood is intractable \citep{tran2017variational, ong2018variational}. Complex variational families have been proposed, e.g.~Gaussian mixtures to account for multi-modality \citep{zobay2014, Miller2016} and Gaussian copulas \citep{Han2016} for flexible multivariate modeling.

VI formulates the problem of approximating a probability density as an optimization problem. To implement the optimization efficiently, it is crucial to obtain an accurate estimate of the gradient when the function to be optimized is intractable but can be unbiasedly estimated.  To this end, the reparameterization (RP) trick \citep{KingmaWelling, rezende2014stochastic} has been useful and much more efficient than the original score function method \citep{REINFORCE}. There is now a large literature applying the RP trick successfully in different settings and recently it has been extended to a wider range of variational approximations \citep{ruiz2016generalized, figurnov2018implicit} and even for non-differentiable models \citep{lee2018reparameterization}. Remarkably, despite the abundance of research utilizing the RP trick, its variance reduction properties are not well studied, apart from a few exceptions, which we review in Section \ref{subsec:related_work}.

\paragraph{General Framework} 

We compare the RP trick to the score function method and show that the former yields more efficient gradient estimators under certain simplifying assumptions. Our first main assumption is that the variational approximation is a mean-field Gaussian density, which is a common modelling choice that has been successfully used in many challenging applications \citep[among others]{KingmaWelling, rezende2014stochastic,kucukelbir2017automatic}. Our second main assumption is that the log-joint density of the model parameters and the data is a quadratic function that varies with the variational mean. We refer to this function as the log-joint density for simplicity. For any general log-joint density, applying this assumption is the same as approximating the true log-joint density with its second-order Taylor series expansion around the variational mean. 

These assumptions allow us to derive expressions for the marginal variances of the gradient estimators under the score function method and RP trick. We then show that the RP gradient estimator is more efficient than the score function estimator since it yields lower marginal variances. This is done by finding a lower bound on the score function marginal variance through applying Rao-Blackwellization. Finally, these expressions are used to understand why and when the RP trick is more efficient. 

\paragraph{Contribution} 

Our contribution is to both prove and understand why the RP trick yields more efficient gradients than the score function method under the simplifying assumptions above. We conclude that:
\begin{itemize}[topsep=0pt,itemsep=0pt,partopsep=0pt, parsep=0pt]
    \item The score function method yields an estimator containing higher order powers of $\theta$ than that of the RP trick, resulting in the score function estimator ``varying" more over its ``sampling region". Section \ref{sec:Interpretations} elaborates further and illustrates this with a simple example.
    \item The marginal variance of each element in both the score function and RP gradient increases with the local ``curvature" of the log-joint density around the variational mean. Furthermore, the marginal variances under the score function method tends to be smaller when the variational mean is close to the true posterior mode. This does not occur under the RP trick.
    \item The marginal variances of the gradient under the score function method increase as the variational scale parameters tend to 0 unlike the RP trick.
    \item Section \ref{sec:observations} discusses other fundamental differences between the gradients.
\end{itemize}

\section{STOCHASTIC GRADIENT VARIATIONAL INFERENCE\label{sec:stochasticgradientvariational}}
\subsection{The variational lower bound}
Let $y = \{y_1, \dots, y_n\}$ denote a dataset with $n$ observations, where $y_i \in \mathcal{X} \subseteq \R^{l}$ for all $i$. Given a model parameterized by $\theta \in \Theta \subseteq \R^k$, with prior density $p(\theta)$, the posterior density is 
\begin{equation}
p(\theta | y) = p(y | \theta)p(\theta)/p(y), \label{eq:posterior_density}
\end{equation}
where $p(y|\theta)$ denotes the model likelihood, $p(y) = \int_{\theta\in\Theta} p(y , \theta) d\theta$ is the marginal likelihood or evidence and $p(y , \theta) = p(y | \theta)p(\theta)$ is the joint density of $y$ and $\theta$. Bayesian inference generally involves computing expectations of functions of $\theta$ with respect to \eqref{eq:posterior_density} which usually does not belong to a known family of densities. 

The goal of VI is to approximate the posterior density in \eqref{eq:posterior_density} by using an appropriate approximating family of variational densities $q(\theta; \lambda)$, where $\lambda = \{\lambda_1, \dots, \lambda_p\}$ are the variational parameters with $\lambda_i \in \Lambda_i \subset \R^{p_{\lambda_i}}$  where $p_{\lambda_i}$ is the number of variational parameters in parameter set $i$ and $p$ is the number of parameter sets in the variational approximation. For example, if $q(\theta;\lambda)$ is Gaussian, then $p = 2$, where $\lambda_1 \in \mathbb{R}^{k}$ is the mean and $\lambda_2 \in \mathbb{R}^{k(k + 1)/2}$ are the unique elements of the covariance matrix. VI finds the optimal $\lambda$ by minimizing the Kullback-Leibler (KL) divergence between the approximation and the true posterior density, 
\begin{align} 
\text{KL}(q(\theta; \lambda) \| p(\theta | y))&=\int_{\theta \in \Theta}q(\theta; \lambda)\log \frac{q(\theta; \lambda)}{p(\theta|y)}d\theta \nonumber\\&=\EE_{q}[\log q(\theta; \lambda)-\log p(\theta|y)], \label{eq:KL}
\end{align}
where $\EE_{q}[\cdot]$ denotes expectation with respect to density $q(\cdot)$.  The KL divergence is non-negative and is zero if and only if $q(\theta; \lambda)= p(\theta|y)$. Computing \eqref{eq:KL} requires evaluating $p(y)$, which is typically intractable. A tractable approach is obtained by maximizing an alternative objective function, which is equivalent to minimizing the KL divergence. We have that
\begin{align}
  \log p(y) & = {\cal L}(\lambda)+\text{KL}(q(\theta; \lambda) \| p(\theta | y)),  \label{logpy}
\end{align}
where 
\begin{align}
  {\cal L}(\lambda) & = \int \log \left(\frac{p(y, \theta)}{q(\theta; \lambda)}\right)q(\theta; \lambda)\;d\theta\nonumber\\
   &= \EE_q[h(\theta) - \log q(\theta; \lambda) ] , \label{lb}
\end{align}
is referred to as the evidence lower bound (ELBO) because $\log p(y) \geq {\cal L}(\lambda)$ and $h(\theta)= \log p(y, \theta)$. Eq.~(\ref{logpy}) shows that minimizing
the KL divergence is equivalent to maximizing the ELBO in (\ref{lb}), which does not require evaluating $p(y)$.

\subsection{Stochastic gradient optimization}

The gradient of the ELBO in \eqref{lb} is rarely available in closed form. Stochastic gradient methods \citep{robbins1951stochastic, bottou10} are useful for optimizing an objective function whose gradient can be unbiasedly estimated. Let $\nabla_\lambda\ELBO(\lambda)$ be the gradient vector of $\ELBO(\lambda)$ in \eqref{lb} with respect to $\lambda$. There are numerous ways to represent this gradient, each one giving a specific estimator: see \cite{roeder2017sticking} for some choices. We use the following representation
\begin{equation}\label{eq:Gradient_raw_form}
\nabla_\lambda \mathcal{L}(\lambda) = \nabla_\lambda \EE_q [h(\theta)] + \nabla_\lambda \mathbb{H}_q[q(\theta; \lambda)],
\end{equation}
where $\mathbb{H}_q[q(\theta; \lambda)] = -\mathbb{E}_q[\log q(\theta; \lambda)] $ is the entropy of $q$ and is analytically solvable when the variational density is Gaussian (Assumption \ref{ass:multivariate_normal}). In the rest of the article, whenever the entropy term appears in an estimator, it is evaluated explicitly. 

Let $\widehat{\nabla_\lambda \ELBO(\lambda)}$ be an unbiased estimator of the gradient which we obtain by Monte Carlo simulation as follows. Suppose that the first term of $\nabla_\lambda \ELBO(\lambda)$ in \eqref{eq:Gradient_raw_form} may be written as an expectation of a function $\Delta(\theta; \lambda)$ with respect to a density $g(\theta; \lambda)$. Then, providing that sampling from $g(\theta; \lambda)$ is possible, an unbiased estimate of $\nabla_\lambda \ELBO(\lambda)$ can be constructed through
\begin{align}
    \widehat{\nabla_\lambda \ELBO(\lambda)} &= \frac{1}{S}\sum_{s=1}^{S}  \Delta(\theta^{(s)}; \lambda) +\nabla_\lambda \mathbb{H}_q[q(\theta; \lambda)], \label{eq:mcgrad} \\  
    &\theta^{(s)} \sim g(\theta; \lambda), \, s = 1, \dots, S. \nonumber
\end{align}
Now, starting from $\lambda=\lambda^{(0)}$, the iteration
\begin{equation}\label{eq:gradupdate}
    \lambda^{(t+1)} = \lambda^{(t)} + \eta_t \circ \widehat{\nabla_\lambda \ELBO(\lambda^{(t)})}
\end{equation}
may be performed until some convergence criteria on $\ELBO(\lambda)$ is met, where the vector $\eta_t$ is a sequence of learning rates and $\circ$ denotes the Hadamard product (element-wise multiplication). Under certain regularity conditions, and when the learning rates satisfy the Robbins-Monro conditions 
\begin{equation*}
   \sum_{t=0}^\infty \eta_t = \infty \quad\text{and}\quad \sum_{t=0}^\infty \eta_t^2 < \infty,
\end{equation*}
the iterates converge to a local optimum \citep{robbins1951stochastic}. Adaptive learning rates are currently popular \citep{duchi2011adaptive, zeiler2012adadelta, kingma2014adam} and we use Adam \citep{kingma2014adam} in our empirical examples, but this choice does not affect our results or conclusions. 

The efficiency of the optimization when iterating \eqref{eq:gradupdate}, i.e.~how fast it converges, depends on how one expresses $\nabla_\lambda\ELBO(\lambda)$; different parametrizations (different $\Delta$ and/or $g$), give rise to different estimators, all of which are unbiased but may have very different variances. For each parameterization, the accuracy of the estimator in \eqref{eq:mcgrad} also depends on the number of Monte Carlo samples $S$. Our article considers three gradient estimators: the RP gradient \citep{KingmaWelling, rezende2014stochastic}, the score function gradient \citep{REINFORCE} and a Rao-Blackwellized version of the score function gradient \citep{ranganath2014black}. The Rao-Blackwellization is used to derive lower bounds for the marginal variances of the score function gradient. Under our assumptions we show that the marginal variances of the RP gradient are less than or equal to the Rao-Blackwellized score function gradient. It trivially follows that the trace of the covariance matrix is smaller for the RP gradient, explaining its superiority over the score function gradient. 

\section{FRAMEWORK\label{sec:Framework}}
\subsection{Structure of the variational approximation}
\begin{assumption}\label{ass:multivariate_normal}
The variational approximation is $q(\theta; \lambda) = \mathcal{N}(\theta|\mu, \Sigma)$, with $\mu = (\mu_1, \dots, \mu_k)^\top$ and $\Sigma = \mathrm{diag}(\exp(2\phi_1), \dots, \exp(2\phi_k))$, $\phi_i = \log(\sigma_i), \sigma_i = \Sigma_{ii}^{1/2}$, where $\mathcal{N}(\cdot | \mu, \Sigma)$ denotes the Gaussian density with mean vector $\mu$ and (diagonal) covariance matrix $\Sigma$.
\end{assumption}
Assumption \ref{ass:multivariate_normal} implies an independence structure known as a mean-field approximation and has been extensively used in conjuction with stochastic gradient methods \citep[among others]{KingmaWelling, rezende2014stochastic,kucukelbir2017automatic}. Under this assumption, the variational density takes the form
\begin{equation}\label{eq:Product_variational}
q(\theta; \lambda) = \prod_{i=1}^k \mathcal{N}(\theta_i|\mu_i, \exp(2\phi_i)),
\end{equation}
with variational parameters $\mu = (\mu_1, \dots, \mu_k)^\top$ and $\phi = (\phi_1, \dots, \phi_k)^\top$, and the vector of all variational parameters is $\lambda = (\mu^\top, \phi^\top)^\top$. There are two reasons we use $\phi_i$ instead of $\sigma_i$: the optimization is easier as it is unrestricted and, moreover, Assumption \ref{ass:CLT} in the next subsection becomes more plausible.
\subsection{Comparing gradient estimators\label{subsec:CompareGradEstimators}}
The gradient of $\mathcal{L}(\lambda)$ is partitioned as
$$\nabla_\lambda \ELBO(\lambda) = (\nabla_{\mu} \ELBO(\lambda)^\top, \nabla_{\phi} \ELBO(\lambda)^\top)^\top,$$
with its estimator
\begin{equation}\label{eq:Decompose_gradientestimators}
\widehat{\nabla_\lambda \ELBO(\lambda)} = \left(\widehat{\nabla_{\mu} \ELBO(\lambda)}^\top , \widehat{\nabla_{\phi} \ELBO(\lambda)}^\top\right)^\top,
\end{equation}
where $\lambda = (\mu^\top, \phi^\top)^\top \in \mathbb{R}^{2k}$ contains all of the variational parameters. The Central Limit Theorem (CLT) motivates the next assumption. Recall that the entropy term $\mathbb{H}_q[q(\theta; \lambda)]$ is assumed known.
\medskip
\begin{assumption}\label{ass:CLT}
Let
\begin{multline}\label{eq:mcgrad_CLT}
    \widehat{\nabla_{\lambda} \ELBO(\lambda)} = \frac{1}{S}\sum_{s=1}^{S} \Delta(\theta^{(s)}; \lambda) + \nabla_\lambda \mathbb{H}_q[q(\theta; \lambda)],  \\
     \theta^{(s)} \overset{\mathrm{iid}}{\sim} g(\theta; \lambda), \quad \theta \in \mathbb{R}^k, 
\end{multline}
where $\Delta : \mathbb{R}^k \to \mathbb{R}^{2k}$ and $g(\theta; \lambda)$ is any density. We assume that for each $j=1,\dots, 2k,$   
$$ \widehat{\nabla_{\lambda} \ELBO(\lambda)_j} \sim \mathcal{N}\left(\nabla_\lambda \ELBO(\lambda)_j , \frac{1}{S}\Var_g\left(\Delta_j(\theta; \lambda)\right)\right),$$
where $\widehat{\nabla_{\lambda} \ELBO(\lambda)_j}$ and $\nabla_\lambda \ELBO(\lambda)_j$ are the $j$-th elements of the corresponding vectors, and $\Delta_j(\theta; \lambda)$ denotes the $j$-th element of $\Delta(\theta; \lambda)$. \end{assumption}
The CLT approximately holds even for small values of $S$ due to independent sampling from $g$. We have found empirically that the transformation $\phi_i = \mathrm{log}(\sigma_i)$, $i = 1, \dots, k$, makes Assumption \ref{ass:CLT} more plausible in practice since it corrects for skewness.

Assumption \ref{ass:CLT} allows us to only consider the marginal variances when comparing unbiased estimators for the $j$-th element obtained with different $\Delta$-functions. \cite{Balles2018dissecting} also consider only the marginal variances when studying the effect of the variability of the stochastic gradient on the Adam optimizer \citep{kingma2014adam}. 

To compare the efficiency of the full (vector) gradient estimator, we follow \cite{miller2017reducing} and consider the trace metric, which is the trace of the estimator covariance matrix, as a scalar measure of variability. This is justified by Assumption \ref{ass:CLT} and allows us to establish analytical results. Under our assumptions, this metric is smaller for the RP gradient compared to the score function gradient. 
Alternative scalar metrics which capture dependencies between gradient components exist. \citet{roeder2017sticking} use the nuclear norm of the estimator covariance matrix. Another metric is the generalized variance \citep{wilks1932certain}, defined as the determinant of the estimator covariance matrix. However, these metrics are analytically intractable under our assumptions. Furthermore, they rely on the multivariate CLT because the covariance matrix is only useful for comparing variability between multivariate Gaussian random variables. For high-dimensional $\lambda$, the multivariate CLT requires a prohibitively large $S$ and is therefore not appropriate in practice. 


\subsection{Gradient estimators\label{subsec:GradientEstimators}}
The RP trick assumes that $\theta \sim q(\theta; \lambda)$ can be written as $\theta=T(z; \lambda)$, $T : \mathbb{R}^k \to \mathbb{R}^{k}$, where $z$ is a random vector (with the same dimension as $\theta$) with density $f(z)$ which does not depend on the variational parameters $\lambda$. This describes a generative model for $\theta$ in terms of the variational parameters. For example, when $q(\theta; \lambda )\sim\mathcal{N}(\mu, \mathrm{diag}(\exp(2\phi)))$, then $T(z; \lambda) = \mu +  \exp(\phi)\circ z$ with $z \sim \mathcal{N}(0,I)$, where $I$ is the $k\times k$ identity matrix and the exponential function is applied element-wise. The gradient of the ELBO under reparameterization becomes
\begin{align}
    \nabla_\lambda\EE_q[h(\theta)] &= \EE_{f}[\nabla_\lambda T(z;\lambda)\nabla_\theta h(\theta)\rvert_{\theta=T(z;\lambda)}], \label{eq:rpgrad0}
\end{align}
where $\Delta^{\mathrm{RP}}(z; \lambda) = \nabla_\lambda T(z;\lambda)\nabla_\theta h(\theta)\rvert_{\theta=T(z;\lambda)}$  and $h:\mathbb{R}^k \to \mathbb{R},$ using RP to emphasize that it is the $\Delta$-function in \eqref{eq:mcgrad} (now a function of $z$) for the RP trick. The gradient of the ELBO under the RP trick is
\begin{multline}\label{eq:ELBOgradrp}
    \nabla_\lambda\ELBO(\lambda)_{\mathrm{RP}} = \EE_f[\nabla_\lambda T(z;\lambda)\nabla_\theta h(\theta)\rvert_{\theta=T(z;\lambda)}] \\  + \nabla_\lambda\mathbb{H}_q[q(\theta; \lambda)],
\end{multline}
and an unbiased estimate is obtained by
\begin{multline}\label{eq:mcgrad_RP}
    \widehat{\nabla_\lambda \ELBO(\lambda)}_{\mathrm{RP}} = \frac{1}{S}\sum_{s=1}^{S}  \Delta^{\mathrm{RP}}(z^{(s)}; \lambda) +\nabla_\lambda \mathbb{H}_q[q(\theta; \lambda)], \\ \quad z^{(s)} \sim f(z), \, s = 1, \dots, S.
\end{multline}

The score function method, also known as the log-derivative trick or the REINFORCE algorithm \citep{REINFORCE}, expresses the gradient of the first term in \eqref{eq:Gradient_raw_form} as
\begin{equation*}
\nabla_\lambda \EE_q[h(\theta)] = \EE_q[h(\theta)\nabla_\lambda \log q(\theta; \lambda)].
\end{equation*}
For this estimator, the $\Delta$-function in \eqref{eq:mcgrad} is $\Delta^{\mathrm{score}}(\theta; \lambda) = h(\theta)\nabla_\lambda \log q(\theta; \lambda)$. The gradient of the ELBO under the score function method is
\begin{multline}\label{eq:ELBOgradscorefunctionmethod}
    \nabla_{\lambda} \ELBO(\lambda)_{\mathrm{score}} = \EE_q[h(\theta)\nabla_\lambda \log q(\theta; \lambda)] \\ + \nabla_\lambda\mathbb{H}_q[q(\theta; \lambda)], 
\end{multline}
and an unbiased estimate is obtained by
\begin{multline}\label{eq:mcgrad_score}
    \widehat{\nabla_\lambda \ELBO(\lambda)}_{\mathrm{score}} = \frac{1}{S}\sum_{s=1}^{S}  \Delta^{\mathrm{score}}(\theta^{(s)}; \lambda)  + \nabla_\lambda \mathbb{H}_q[q(\theta; \lambda)], \\ \theta^{(s)} \sim q(\theta; \lambda), \, s = 1, \dots, S.
\end{multline}

We use a Rao-Blackwellized score function gradient estimator introduced by \citet{ranganath2014black} to find a lower bound for the marginal variances of the score function estimator and show that the corresponding variances under the RP gradient are smaller. To implement the Rao-Blackwellization (RB), suppose that the variational approximation satisfies Assumption
\ref{ass:multivariate_normal} and define $h_{-i}(\theta)$ to be $h(\theta)$ with any elements not containing $\theta_i$ removed. Furthermore, denote the Markov blanket of the $i$-th parameter as $\theta_{(i)}$, see Section \ref{app:Proofs} of the supplementary material for details. The gradient in (\ref{eq:rpgrad0}) may be written as an iterated conditional expectation, which for $i=1,\dots,k$, simplifies to
\begin{multline}\label{eq:ELBOgradRB}
\nabla_{(\mu_{i}, \phi_{i})}\E_q[h(\theta)] = \E_{q_{(i)}}[h_{-i}(\theta_{(i)}) \\ \nabla_{(\mu_{i}, \phi_{i})} \log q(\theta_{i}; \mu_{i}, \phi_{i})],
\end{multline}
where $q_{(i)}$ is the density of $\theta_{(i)}$. Hence, we define $\Delta^\text{RB}(\theta_{(i)}; \lambda) = h_{-i}(\theta_{(i)})\nabla_{(\mu_{i}, \phi_{i})} \log q(\theta_{i}; \mu_{i}, \phi_{i})$ and form the Rao-Blackwellized gradient estimator for the $i$-th component as
\begin{multline}\label{eq:RBMCgrad}
\reallywidehat{\nabla_{(\mu_{i}, \phi_{i})}\ELBO(\lambda)}_\text{RB} = \frac{1}{S}\sum_{s=1}^{S}  \Delta^\text{RB}(\theta^{(s)}; \lambda)  + \nabla_{(\mu_{i}, \phi_{i})}\mathbb{H}_{q_{(i)}}[\\ q(\theta; \mu_{i}, \phi_{i})], \quad \theta^{(s)}\sim q_{(i)}(\theta ; \mu_{i}, \phi_{i}), \, s = 1, \dots, S.
\end{multline}
The full estimator, i.e.~$\widehat{\nabla_\lambda\ELBO(\lambda)}_\text{RB}$, is obtained by merging \eqref{eq:RBMCgrad} for $i = 1, \dots, k$ and ordering them as $\lambda =(\mu^\top, \phi^\top)^\top$. For details and a full derivation, see \citet{ranganath2014black} and Section \ref{app:Proofs} of the supplementary material.

\begin{small}
\centering
\captionsetup{width=\linewidth}
\captionof{table}{$\Var_{q}\left[\Delta^\mathrm{score}(\theta;\lambda)\right]$ and $\Var_{f}\left[\Delta^\mathrm{RP}(z;\lambda)\right]$ estimated using $S=10,000$ samples. The approximations deteriorate as $\sigma = \exp(\phi)$ increases. True refers to the using the true log-joint density, and approx. refers to replacing $h(\theta)$ with the quadratic approximation.}\label{tab:approx_variance_diff}%
\begin{tabular}{lcccc} \hline
\backslashbox[24mm]{$\widehat{\nabla\mathcal{L}(\lambda)}$}{$\sigma$}
   & (0.1, 0.1) & (0.5, 0.5) & (1, 1) & (2,2)\\ \hline
  Score (true)& 32,459 & 1,648  & 439 & 229  \\
  Score (approx.) & 32,459 & 1,659  & 473 & 369 \\ 
  RP (true) & 0.06 & 1.40 & 3.56 & 7.76 \\
  RP (approx.) & 0.06 & 1.64 & 5.69 & 24.05 \\\hline
\end{tabular}
\end{small}

\subsection{Structure of the log-joint density\label{subsec:Structure_log_joint}}
We now present an assumption that allows us to (i) obtain analytical expressions for the marginal variances of the score function and RP gradient estimators and (ii) understand how the RP trick reduces the variance. 

\begin{assumption}\label{ass:LocallyQuadratic}
Let $\mu = (\mu_1, \dots, \mu_k)^\top$ be the variational mean and suppose that the log-joint density $h(\theta)= \log p (y, \theta)$ is given by
\begin{equation}\label{eq:quadratic_log_joint}
h(\theta) = C + G(\mu)^\top (\theta-\mu) + \frac{1}{2} (\theta - \mu)^\top  H(\mu) (\theta - \mu) ,
\end{equation}
where $C$ is a constant, $G(\mu)$ is a vector whose entries are functions of $\mu$ and $H(\mu)$ is a symmetric matrix. 
\end{assumption} 

We refer to Assumption \ref{ass:LocallyQuadratic} as the quadratic assumption on the log-joint density. We can liken this to a second-order Taylor series expansion of any general log-joint density around the variational mean. In this case, $G(\mu) = \nabla_\theta h(\mu)$ and $H(\mu)$ is the hessian of $h(\theta)$ evaluated at $\mu$. 

The plausibility of Assumption \ref{ass:LocallyQuadratic} depends on how far the sampled values of $\theta$ are from $\mu$ when evaluating the Monte Carlo gradients and to what degree the true $h(\theta)$ is quadratic in this region. We would expect that as $\phi$ increases, more samples lie in a region where the approximation is poor and so the corresponding estimates of the marginal variances will deteriorate.

We now introduce a simple Bayesian logistic regression model as a running example for the rest of the paper to illustrate our assumptions and findings. We generate $n=10$ observations from a logistic regression model, with input $x\in\R$, response $y\in\{0,1\}$ and $p(y |x,\theta) = p(x)^y(1-p(x))^{1-y}$, where $p(x) = 1 / (1 + e^{\theta_1 + \theta_2 x})$. Furthermore, we set a $\mathcal{N}(0, \sigma_0^2 I)$ prior on $\theta$ where $\sigma_0 = 5$ and apply a mean-field Gaussian variational approximation $q(\theta; \mu, \phi)$. Table \ref{tab:approx_variance_diff} illustrates how increasing $\phi$ causes the approximations to the marginal variances deteriorate in this example.


\subsection{Results \label{sec:ResultsAndInterpretations}}
The following proposition gives the marginal variances for the RP gradient and shows that they are smaller or equal to those of the score function gradient. Section \ref{app:Proofs} of the supplementary material provides a proof.

\begin{figure}
    \begin{subfigure}[htp]{0.45\textwidth}
      \includegraphics[width=\textwidth]{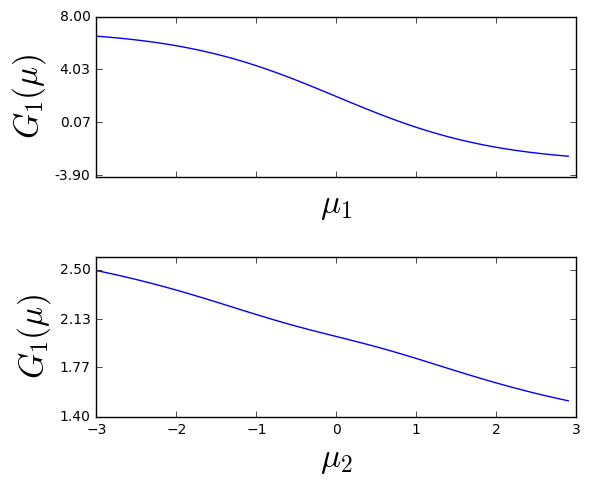}
    \end{subfigure}
    \begin{subfigure}[htp]{0.45\textwidth}
      \includegraphics[width=\textwidth]{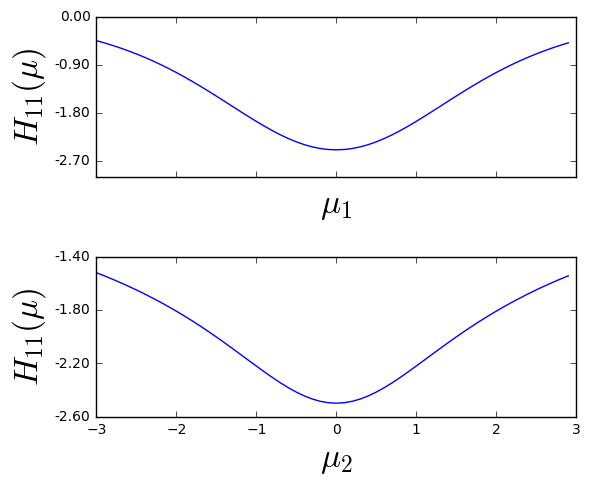}
    \end{subfigure}
    \begin{subfigure}[htp]{0.45\textwidth}
      \includegraphics[width=\textwidth]{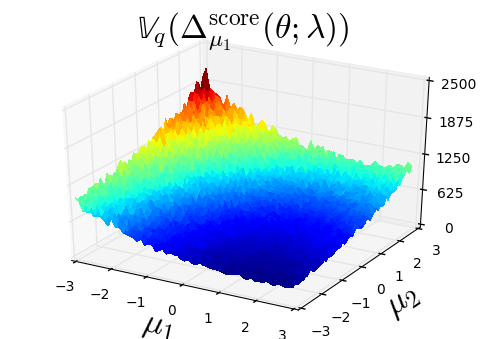}
    \end{subfigure}
    \begin{subfigure}[htp]{0.45\textwidth}
      \includegraphics[width=\textwidth]{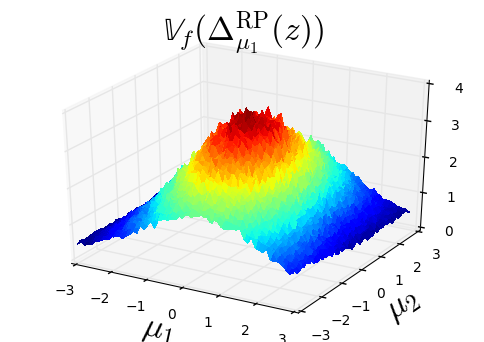}
    \end{subfigure}
     \captionsetup{width=\linewidth}
    \caption{\textbf{Top row:} Cross section of $G_1(\mu)$ (left) and $H_{11}(\mu)$ (right) for $\mu_2=0$ (top) and $\mu_1=0$ (bottom). \textbf{Bottom row:} Simulation estimates of $\mu_1$ gradient marginal variances ($S=1,000$) with $\sigma_i=1$ for $i=1,2$. The score function variance increases with $G_1(\mu)$ whereas the RP variance depends more on $H(\mu)$.}\label{fig:observations_var_surface_mu}
  \end{figure} 

\begin{proposition}\label{lem:marginal_variances}
Suppose that Assumptions \ref{ass:multivariate_normal}-\ref{ass:LocallyQuadratic} hold and let $T(z; \lambda) = \mu + \sigma \circ z$, where $\mu = (\mu_1, \dots, \mu_k)^\top$, $\sigma = (\sigma_1, \dots,  \sigma_k)^\top$, $\sigma_i = \exp(\phi_i)$ and $z = (z_1, \dots, z_k)^\top$ with $z_i \sim \mathcal{N}(0, 1)$. Then, for $i = 1,\dots, k$,
\begin{enumerate}[topsep=0pt, label={\emph{(\roman*)}}]
\item \begin{multline}\label{eq:Variance_mu_score}
    \Var_q\left(\Delta^\mathrm{score}_{\mu_i}(\theta; \lambda)\right) = \frac{1}{\sigma_i^2}(C^2 + C\mathrm{diag}(H(\mu)^{2})^\top \sigma^2 +\\ 2C\sigma_i^2H_{ii}(\mu) G(\mu)^{2\top} \sigma^2  + \sigma_i^2G_i(\mu)^2 ) + \\Q(H(\mu), \sigma) 
\end{multline}
\begin{multline}\label{eq:Variance_phi_score}
    \Var_q\left(\Delta^\mathrm{score}_{\phi_i}(\theta; \lambda)\right) = 3C^2 + CH_i(\mu)^\top \sigma^2 \\+ 4C\sigma^2 H_{ii}(\mu) + 
    3( G(\mu)^{2\top} \sigma^2 + \\4G_i(\mu)^2\sigma^2) + R(H(\mu), \sigma),
\end{multline}
where $C$ is a constant independent of $\lambda$ and $Q(H(\mu), \sigma)$ and $R(H(\mu), \sigma)$ are second order function of elements of $H(\mu)$ and $\sigma$.
\item 
\begin{align}
    \Var_f\left(\Delta^\mathrm{RP}_{\mu_i}(z; \lambda)\right) = & H_{i}(\mu)^{2\top} \sigma^2 \label{eq:Variance_mu}
\end{align}
\begin{multline}
\Var_f\left(\Delta^\mathrm{RP}_{\phi_i}(z; \lambda)\right)  =  \sigma_i^2\Bigl( H_{i}(\mu)^{2\top} \sigma^2 + H_{ii}(\mu)^2 \sigma_i^2 \\ + G_i(\mu)^2 \Bigr)\label{eq:Variance_phi},
\end{multline}
\item $$\Var_f\left(\Delta^\mathrm{RP}_{\mu_i}(z; \lambda)\right) \leq \Var_q\left(\Delta^\mathrm{score}_{\mu_i}(\theta; \lambda)\right)$$ and $$\Var_f\left(\Delta^\mathrm{RP}_{\phi_i}(z; \lambda)\right) \leq \Var_q\left(\Delta^\mathrm{score}_{\phi_i}(\theta; \lambda)\right).$$
\end{enumerate}
\end{proposition}

Corollary \ref{cor:trace_metrics} shows that the trace of the covariance matrix of the RP gradient is smaller than that of the score function gradient. 
\begin{corollary}\label{cor:trace_metrics}
Suppose Assumptions 1--3 hold and define $$\widehat{\nabla_\lambda \ELBO(\lambda)}_{\mathrm{RP}} \text{ and } \widehat{\nabla_\lambda \ELBO(\lambda)}_{\mathrm{score}}$$
as in Section \ref{subsec:GradientEstimators}. Then, 
\begin{multline*}
\mathrm{tr}\left(\Cov_f\left(\widehat{\nabla_\lambda \ELBO(\lambda)}_{\mathrm{RP}}\right)\right) \leq  \mathrm{tr}\left(\Cov_q\left(\widehat{\nabla_\lambda \ELBO(\lambda)}_{\mathrm{score}}\right)\right).
\end{multline*}
\end{corollary}

 \begin{figure}
    \begin{subfigure}[htp]{0.45\textwidth}
      \includegraphics[width=\textwidth]{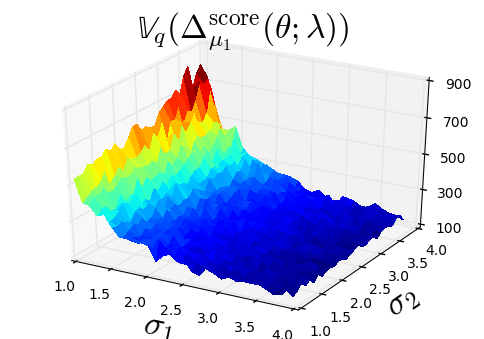}
    \end{subfigure}
    \begin{subfigure}[htp]{0.45\textwidth}
      \includegraphics[width=\textwidth]{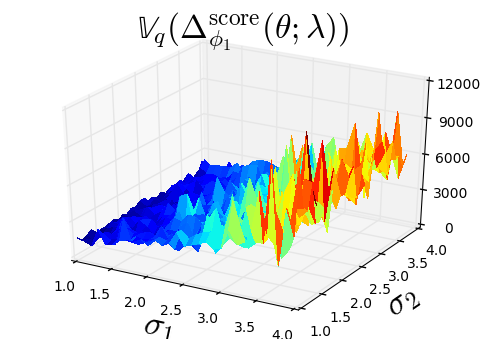}
    \end{subfigure}
    \begin{subfigure}[htp]{0.45\textwidth}
      \includegraphics[width=\textwidth]{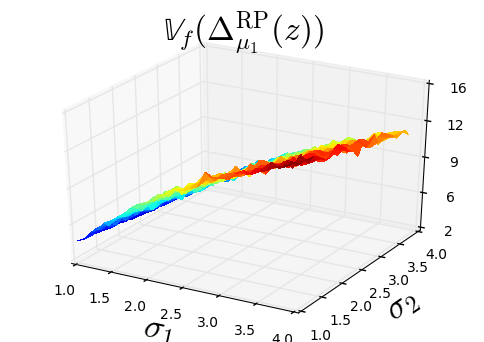}
    \end{subfigure}
    \begin{subfigure}[htp]{0.45\textwidth}
      \includegraphics[width=\textwidth]{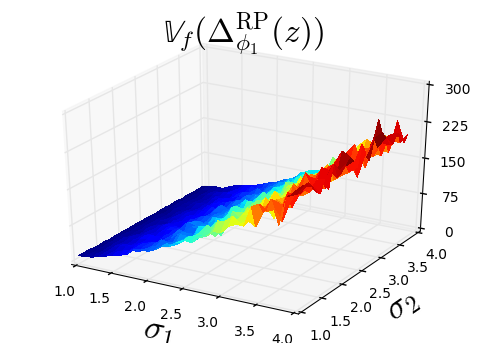}
    \end{subfigure}
     \captionsetup{width=\linewidth}
    \caption{\textbf{Top row:} Simulation estimates of score function marginal variances ($S=1,000$) with $\mu_i=0$ for $i=1,2$. $\Var_q(\Delta^{\mathrm{score}}_{\mu_1}(\theta; \lambda))$ increases as $\sigma_1\rightarrow 0$, but not when $\sigma_2\rightarrow 0$. \textbf{Bottom row:} As per top row but for RP. No deterioration occurs as $\sigma_1\rightarrow 0$.}\label{fig:observations_var_surface_sigma}
  \end{figure}
  
  \begin{figure}
    \begin{subfigure}[htp]{0.45\textwidth}
        \includegraphics[width=\textwidth]{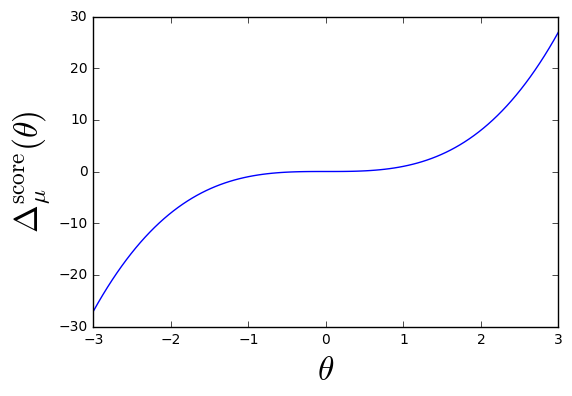}
    \end{subfigure}
    \begin{subfigure}[htp]{0.45\textwidth}
        \includegraphics[width=\textwidth]{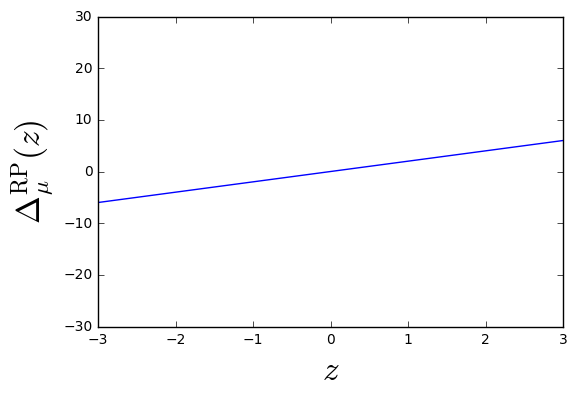}
    \end{subfigure} 
    \begin{subfigure}[htp]{0.45\textwidth}
        \includegraphics[width=\textwidth]{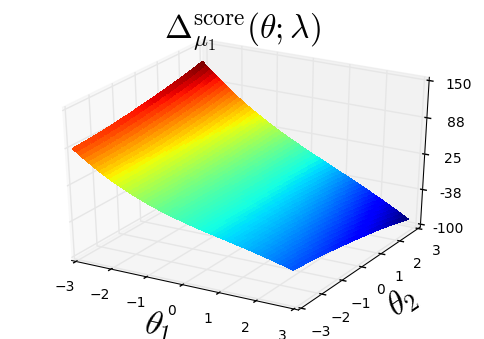}
    \end{subfigure}
    \begin{subfigure}[htp]{0.45\textwidth}
        \includegraphics[width=\textwidth]{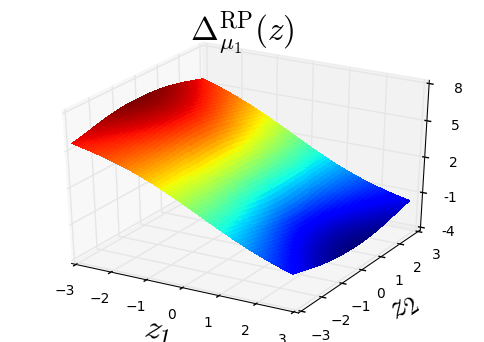}
    \end{subfigure}
    \captionsetup{width=\linewidth}
    \caption{\textbf{Top row:} $\Delta$-functions for $h(\theta)=\theta^2$ and $q(\theta)\sim \mathcal{N}(0,1)$. Note $\Delta^\mathrm{RP}(z; \lambda)$ varies less than $\Delta^\mathrm{score}(\theta;\lambda)$ over the sampling region. \textbf{Bottom row:} $\Delta$-functions for the logistic regression example with $\mu = (0,0)$ and $\sigma = (1, 1)$. Again, notice the higher variation of $\Delta^\mathrm{score}(\theta;\lambda)$ over the sampling region.} 
    \label{fig:insights}
\end{figure}

\subsection{Observations on results}\label{sec:observations}

\begin{figure*}
\begin{tabular}{ccc}
     \includegraphics[width=0.3\textwidth]{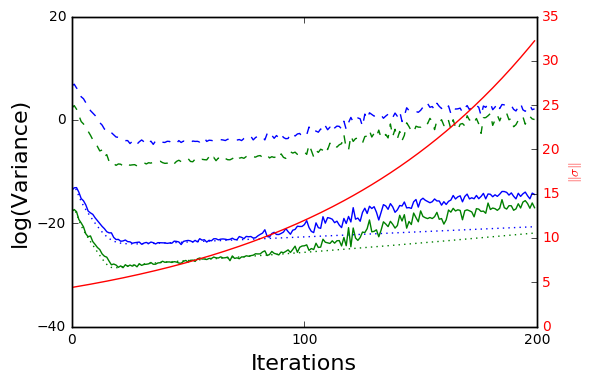} &   \includegraphics[width=0.3\textwidth]{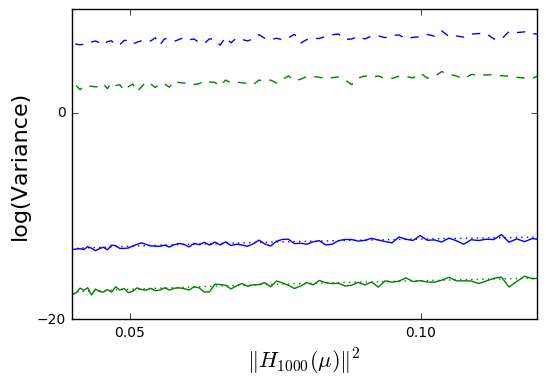}
     & \includegraphics[width=0.3\textwidth]{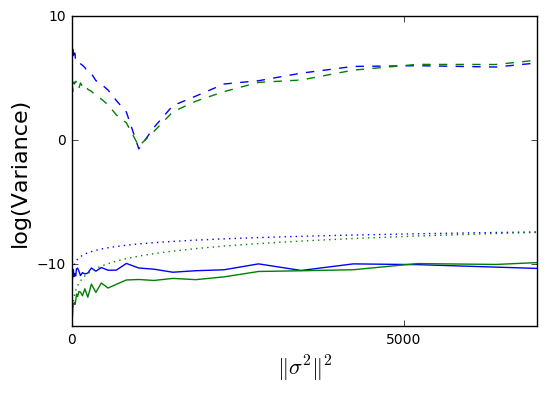}
\end{tabular}
\captionsetup{width=\linewidth}
\caption{Bayesian multinomial logistic regression example under Experiments 1--3 (from left to right), see Section \ref{subsec:Ex1}. Legend: $\mu_{1000}$ (blue), $\phi_{1000}$ (green), Score (dashed), RP (solid), approximations \eqref{eq:Variance_mu} and \eqref{eq:Variance_phi} (dotted). \textbf{Left:} Poor approximations at iteration $100$ are due to high values in $\sigma$. \textbf{Middle:} The variance increases with $\|H_{1000}(\mu)\|$. \textbf{Right:} The variance increases with $\|\sigma\|^2$ despite the poor accuracy of the approximation.}\label{fig:logistic}
\end{figure*}

The expressions derived in Section \ref{sec:ResultsAndInterpretations} yield some intuition behind the differences in marginal variances between the score function and RP gradients. Firstly, the marginal variances of the score function gradient given in \eqref{eq:Variance_mu_score} and \eqref{eq:Variance_phi_score} depend on $G(\mu)$, meaning we would expect the marginal variance to be lowest when $\mu$ is near the true posterior mode where the gradient is 0. In contrast, the RP gradient marginal variances given in \eqref{eq:Variance_mu} and \eqref{eq:Variance_phi} have very little dependence on $G(\mu)$. Furthermore, \eqref{eq:Variance_mu_score} contains a $1 / \sigma_i^2$, which implies that $\Var_q\left(\Delta^\mathrm{score}_{\mu_i}(\theta; \lambda)\right)\rightarrow \infty$ as $\sigma_i \rightarrow 0$. Interestingly, this is not the case for $\Var_q\left(\Delta^\mathrm{score}_{\phi_i}(\theta; \lambda)\right)$ or the RP gradients. Finally, the $\mu_i$ and $\phi_i$ components of the RP gradient only contain gradient component $i$ and row $i$ hessian terms. In contrast, the score function gradient contains all gradient and hessian components. This is due to the RP gradient taking the gradient of the log-joint density, causing all terms not containing $\theta_i$ to vanish. These observations imply that the score function gradient estimator behaves in a fundamentally different way to the RP gradient estimator. Figures \ref{fig:observations_var_surface_mu} and \ref{fig:observations_var_surface_sigma} illustrate this for the logistic regression example presented in Section \ref{subsec:Structure_log_joint}.

\subsection{Insights on the reparameterization trick \label{sec:Interpretations}}
  
Some papers in the literature explain the success of the RP trick as due to its efficient use of gradient information from the log-joint density \citep{titsias2014doubly, TanNott, QuirozNottKohn} without elaborating further. 

We argue that since the RP trick allows us to take the gradient of the log-joint density $h(\theta)$ with respect to $\theta$ when constructing an estimator, it yields an estimator containing lower order terms with respect to $\theta = T(z; \lambda)$ compared to the score function method. Specifically, $\Delta_{\lambda_i}^\mathrm{score}(\theta; \lambda)$ contains higher orders of $\theta$ whereas $\Delta_{\lambda_i}^\mathrm{RP}(z; \lambda)$ contains lower orders of $z$. Let $B_q\subset \Theta$ be a compact subset of $\Theta$ that contains a large proportion of the samples from $q$ used to evaluate the Monte Carlo estimate of the gradient. We refer to this as the ``sampling region" of $q$. Similarly, $B_f$ refers to the sampling region of $f$ for the RP gradient. For example, if $q(\theta;\lambda)= \mathcal{N}(0,2)$ then $B_q=[-6,6]$ and $B_f=[-3,3]$ are appropriate since 99.7\% of the samples lie in these intervals. The reason why the score function gradient tends to have higher variance is because the image of $B_q$ under $\Delta_{\lambda_i}^\mathrm{score}(\theta; \lambda)$ tends to have a larger range compared to the image of $B_f$ under $\Delta_{\lambda_i}^\mathrm{RP}(z; \lambda)$. We call this having a ``higher variation" in the sampling region of the estimator.

To illustrate, suppose $h(\theta)=\theta^2$ and $q(\theta ; \mu)\sim \mathcal{N}(\mu, 1)$. We can use \eqref{eq:ELBOgradscorefunctionmethod} to show that $\Delta^{\mathrm{score}}_\mu(\theta; \lambda) = \theta^3 - \theta^2\mu$, which contains a third order power of $\theta$. From this, $\Var_q(\Delta^{\mathrm{score}}_\mu(\theta; \lambda)) = \mu^4 + 14\mu^2 + 15$. In contrast, the RP gradient estimator is given by $\Delta^{\mathrm{RP}}_\mu(z; \lambda) = 2(\mu + z)$ hence $\Var_f(\Delta^{\mathrm{RP}}_\mu(z; \lambda))=4$. We see a large difference in variance that appears to be driven by the fact that the RP gradient estimator's leading term is at least two orders lower than that of the score function gradient estimator. Consequently, the score function estimator has higher variation over its sampling region compared to the RP gradient estimator. Figure \ref{fig:insights} illustrates this for the example above, as well as for the logistic regression example discussed in Section \ref{subsec:Structure_log_joint}. Note that these observations hold for the gradient with respect to $\phi$ as well and readily extends to the multivariate case. Despite many log-joint density functions not being polynomials, we can find a reasonable polynomial approximation over the sampling region using the Stone-Weierstrass theorem \citep{stone1948generalized}. 

\subsection{Related work\label{subsec:related_work}}

\citet{fan2015fast} show that if a function $g:\R^k \rightarrow \R$ is Lipschitz continuous with constant $L$, and $z \sim \mathcal{N}(0,I_k)$, then $\Var[g(z)] \leq L^2$. In addition, they claim that in practice the variance is highly sensitive to $L$. This is similar to the intuition we develop since $L$ tends to give a rough indication of the variation of $g$ which drives the variance. The limitation of using the Lipschitz constant is that even for basic models such as Bayesian linear regression, the log-joint density is not Lipschitz continuous and so these results are not immediately useful. In our work, we apply a more specific simplifying assumption to the log-joint density instead, which allows us to look at specific properties around the variance reduction of the RP gradient in a region where the function is locally quadratic.

\citet[Chapter~3.1.2]{yurinphd} shows that given a univariate $\theta \sim \mathcal{N}(\mu,\sigma^2)$ and assuming certain conditions on $h$ hold, the RP gradient estimator has smaller marginal variances than the corresponding estimator under the score function method. While \citet[Chapter~3.1.2]{yurinphd} defines a set of conditions and proves that the RP gradient has lower marginal variance given these conditions, limited insight is provided around when the RP trick works well in practice. The results are also restricted to a univariate posterior. We tackle this problem in the multivariate case, and offer a set of simplifying assumptions that are reasonable for certain classes of models. Furthermore, we discuss the intuition behind the drivers of the variance of the gradient estimators and why the RP gradient is more efficient than the score function gradient.

Finally, we note that there are no guarantees that the RP trick is more efficient in the general case. A counterexample corresponding to a highly multimodal log-joint density ($h(\theta)=\sin(10\theta)$) is given in \citet[Chapter~3.1.2]{yurinphd}. This highlights the fact that we need to make reasonable simplifying assumptions on $h$ to be able to theoretically conclude that the RP gradient is more efficient than the score function gradient. 

\section{EXAMPLES\label{sec:empirical_examples}}

\begin{figure}
    \begin{subfigure}[h!]{0.45\textwidth}
        \includegraphics[width=\textwidth]{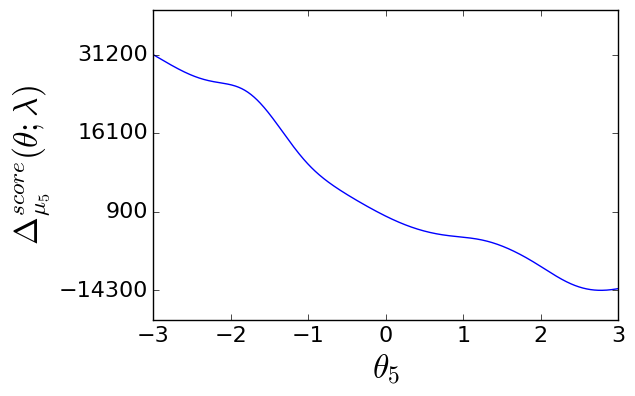}
    \end{subfigure}
    \begin{subfigure}[h!]{0.45\textwidth}
        \includegraphics[width=\textwidth]{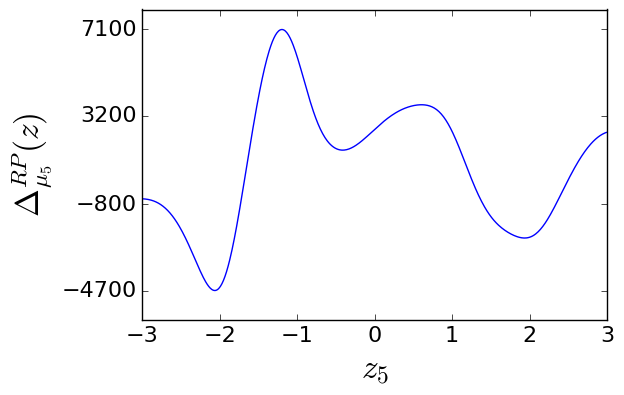}
    \end{subfigure}
    \begin{subfigure}[h!]{0.45\textwidth}
        \includegraphics[width=\textwidth]{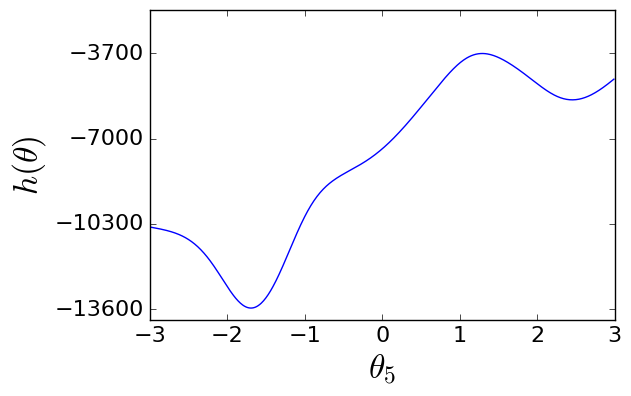}
    \end{subfigure}
    \captionsetup{width=\linewidth}
    \caption{\textbf{Top row:} Cross section of $\Delta_{\mu_5}$ functions for the Bayesian Neural Network model with $q(\theta;\lambda) \sim \mathcal{N}(0,I)$. The RP estimator varies less over its sampling region. Simulations ($S=10,000$) yield $\Var_q(\Delta_{\mu_5}^{\mathrm{score}}(\theta; \lambda)) = 2.53e11$ and $\Var_f(\Delta_{\mu_5}^{\mathrm{RP}}(z; \lambda))=6.60e7$. \textbf{Bottom row:} Cross section of $h(\theta)$. The quadratic assumption is clearly inappropriate here.}\label{fig:BNN_cross_section_5}
\end{figure}

This section studies whether our results and insights from Sections \ref{sec:ResultsAndInterpretations}, \ref{sec:observations} and \ref{sec:Interpretations} derived under the quadratic assumption of the log-joint density are useful in cases where the assumption does not reasonably hold. We show that our expressions for the marginal variances in Section \ref{sec:ResultsAndInterpretations} capture the behaviour of the marginal variances of a high-dimensional multinomial logistic regression model. Furthermore, we show that our intuition regarding the difference in variation of the estimators over the sampling region explains the variance reduction properties of the RP gradient for a simple two layer Bayesian neural network model where we expect the quadratic assumption would not hold. We apply a mean-field Gaussian variational approximation in both examples. 

\subsection{Bayesian multinomial logistic regression}\label{subsec:Ex1}
The MNIST database of handwritten digits \citep{lecun1998gradient} contains 60,000 training observations and 10,000 test observations of $28 \times 28$ images with 10 prediction classes. We fit a Bayesian multinomial logistic (or softmax) regression model for classification with a $\mathcal{N}(0, \sigma_0^2 I)$ prior over the regression coefficients with $\sigma_0 = 40$. The elements of the score function and RP gradient estimators corresponding to parameters $\mu_{1000}$ and $\phi_{1000}$ were analyzed by conducting three experiments. In each case we evaluated the log of the marginal variance for each element.

\paragraph{Experiment 1} 

We ran the optimization for $200$ iterations with $\sigma=\exp(\phi)$ initialized with very small values and observed that all elements of $\sigma$ gradually increased due to the high dimensionality of the posterior relative to the number of observations. 

\paragraph{Experiment 2}  

We held $\phi$ fixed and increased the value of $\mu_i$ while fixing $\mu_j$ for $j\neq i$. This had the effect of varying elements of $H_i(\mu)$. We expect from \eqref{eq:Variance_mu} and \eqref{eq:Variance_phi} that the marginal variance of the gradient for both $\mu_i$ and $\phi_i$ will increase with $\|H_i(\mu)\|^2$ where $\|\cdot\|$ is the Euclidean norm of the corresponding vector. 

\paragraph{Experiment 3}  

We held $\mu$ (and therefore $H_i(\mu)$) fixed and increased $\phi_j$ for all $j$. This was designed to measure the effect of $\phi$ on the marginal variance.  

Figure \ref{fig:logistic} shows the results of Experiments 1--3. When $\sigma$ is small the quadratic assumption yields reasonable estimates for the marginal variances of the RP gradient, but it deteriorates as $\sigma$ increases. In addition, the marginal variance of the score function and RP gradient clearly increases with both $\|H_i(\mu)\|^2$ and $\|\sigma^2\|^2$. Remarkably, this is consistent with both (\ref{eq:Variance_mu}) and (\ref{eq:Variance_phi}), despite these formulas yielding poor estimates of the true marginal variances. 

\subsection{Bayesian neural network}
We follow \citet{duvenaud2015black} and apply a simple Bayesian neural network on 40 simulated observations. The density of observation $y_i$ given input $x_i\in\R$ and neural network weights $\mathbf{w}$ is $p(y_i | \mathbf{w}, x_i, \sigma_\mathrm{err}^2) = \mathcal{N}(y_i | \text{NN}(x_i; \mathbf{w}), \sigma_\mathrm{err}^2)$, where $\text{NN}(x_i; \mathbf{w})$ is a neural network with two hidden layers of size 20 with tanh activations and $\sigma_\mathrm{err}^2=1$. A $\mathcal{N}(0, \sigma_0^2 I)$ prior is set over $\mathbf{w}$ where $\sigma_0 = 40$. Figure \ref{fig:BNN_cross_section_5} illustrates the highly non-quadratic properties of the log-joint density of a neural network. Nevertheless, the variance of the gradient estimators mainly depends on the variation of the estimator over its sampling region.

\section{CONCLUSION AND FUTURE RESEARCH\label{sec:conclusions}}
We have studied the variance reduction properties of the reparameterization trick under certain simplifying assumptions. We argue that its success depends on the fact that it generally results in an expression that has lower variation over the sampling region of the variational distribution compared to the score function method. Finally, we showed that our conclusions in Sections \ref{sec:observations} and \ref{sec:Interpretations} are useful in describing cases where our assumptions are not perfectly satisfied.

Future extensions include relaxing the mean-field assumption by considering more flexible covariance structures as in \citet{TanNott, ONS, QuirozNottKohn}. Variational families other than the Gaussian density may also be considered, for example a mean-field approximation with a mixture of normal and Gamma components like in \citet{ranganath2014black}. Finally, alternative scalar measures of variability such as the ones discussed in Section \ref{subsec:CompareGradEstimators} can be employed to assess the efficiency of the gradient estimators.

\newpage
\section*{Acknowledgements}
The authors were supported by the Australian Centre of Excellence in Mathematical and Statistical Frontiers (ACEMS, grant CE140100049). SAS is also supported by the Australia Research Council Discovery Projects Scheme (grant DP160102544).

\bibliographystyle{apalike}
\addcontentsline{toc}{section}{\refname}\bibliography{ref} 

\setcounter{equation}{0} 
\renewcommand{\theequation}{A\arabic{equation}}
\renewcommand{\thelemma}{A\arabic{lemma}}
\renewcommand{\thedefinition}{A\arabic{definition}}
\renewcommand{\theremark}{A\arabic{remark}}

\begin{appendices}
\section{PROOFS}
\label{app:Proofs}
\begin{proof} To prove (i), let $h(\theta)$ follow \eqref{eq:quadratic_log_joint}. Denote the $i$-th component of $G(\mu)$ by $G_i(\mu)$ and similarly, the $(i,j)$-th component of $H(\mu)$ by $H_{ij}(\mu)$. We can now write
    \begin{align}
        h(\theta) &= C + G(\mu)^\top(\theta - \mu) + \frac{1}{2} (\theta - \mu)^\top  H(\mu) (\theta - \mu) \nonumber\\
        &= C + \sum_{m=1}^k G_m(\mu) (\theta_m-\mu_m)  \nonumber \\  
        & + \frac{1}{2} \sum_{m=1}^k\sum_{n=1}^k (\theta_m-\mu_m)H_{mn}(\mu)(\theta_n-\mu_n). \label{eq:Proof_h_expand}
    \end{align}
    
    Furthermore, let $\sigma_i = e^{\phi_i}$. Given the mean-field Gaussian structure on our variational approximation which assumes that $\theta_i\sim\mathcal{N}(\mu_i, \sigma_i^2)$, we can use standard expressions for a normal density to show that
    \begin{multline}
        \nabla_{(\mu_{i}, \phi_{i})} \log q(\theta_{i}; \mu_{i}, \phi_{i}) = \Bigl( \frac{\theta_i - \mu_i}{\sigma_i ^ 2}, \\ -1 + \frac{(\theta_i-\mu_i)^2}{\sigma_i^2} \Bigr)^\top.\label{eq:Proof_q_grad}
    \end{multline}
    Combining \eqref{eq:Proof_h_expand} and \eqref{eq:Proof_q_grad}, we can now evaluate \eqref{eq:ELBOgradscorefunctionmethod} to be
    \begin{align*}
       \Delta^\mathrm{score}_{\mu_i}(\theta; \lambda) &=  h(\theta)\frac{\partial}{\partial\mu_{i}} \log q(\theta_{i}; \mu_{i}, \phi_{i})\\
       &= h(\theta)\frac{\theta_i - \mu_i}{\sigma_i ^ 2}
    \end{align*}
     and similarly,
    \begin{align*}
        \Delta^\mathrm{score}_{\phi_i}(\theta; \lambda) &=  h(\theta)\frac{\partial}{\partial\phi_{i}} \log q(\theta_{i}; \mu_{i}, \phi_{i})\\
       &= h(\theta) \left(-1 + \frac{(\theta_i-\mu_i)^2}{\sigma_i^2}\right).
    \end{align*}
    
    Finally, using the independence between $\theta_i$ and $\theta_j$ for $i\neq j$ from the mean-field assumption, the standard identity for the variance ($\Var X = \EE X^2 - (\EE X)^2$ for any random variable $X$) and expressions for the moments of normal random variables, yield the results given by \eqref{eq:Variance_mu_score} and \eqref{eq:Variance_phi_score}.
    
    To prove (ii), we find an analytical form for \eqref{eq:rpgrad0} given our assumptions. To begin, again use \eqref{eq:Proof_h_expand} for the log-joint density and find the gradient of this with respect to $\theta$, which gives
    \begin{equation}
        \nabla_\theta h(\theta) = \nabla_\theta h(\mu) + H(\mu)(\theta-\mu)\in \R^{k}.\label{eq:Proof_gradh}
    \end{equation}
    Furthermore, since $\theta = T(z;\lambda) = \mu + \exp(\phi)\circ z \in \R^k$, we can take the gradient with respect to $\mu$ and $\phi$ to show that
    \begin{equation}
        \nabla_\lambda T(z;\lambda) = (I_k, \mathrm{diag}(\exp(\phi)\circ z))^\top \in \R^{2k\times k} \label{eq:Proof_T_grad}.
    \end{equation}
    \eqref{eq:Proof_gradh} and \eqref{eq:Proof_T_grad} can now be combined to find an expression for \eqref{eq:rpgrad0}, and we can take the $i$-th and $2i$-th component of this resulting vector as the estimators for the $\mu_i$ and $\phi_i$ components of the gradient. We can perform standard matrix operations on the above and use $\theta = T(z;\lambda)$ to show that
    \begin{align}
        \Delta^\mathrm{RP}_{\mu_i}(z) & = \frac{\partial}{\partial \theta_i} h(\mu) + \sum_{m=1}^k H_{im}\exp(\phi_i)z_i\label{eq:Proof_grad_RP_mu}\\
        \Delta^\mathrm{RP}_{\phi_i}(z) & = \Delta^\mathrm{RP}_{\mu_i}(z) \exp(\phi_i)z_i.\label{eq:Proof_grad_RP_phi}
    \end{align}
    \eqref{eq:Proof_grad_RP_mu} is a linear transformation of $z$, hence evaluating the variance only requires us to evaluate up to the second moment of a normal distribution. For \eqref{eq:Proof_grad_RP_phi}, we have a quadratic function of $z$ and need to evaluate up the fourth moment of a normal distribution to evaluate the variance. In contrast, the score function method requires us to evaluate up to the $6$-th and $8$-th moments, respectively. Expanding \eqref{eq:Proof_grad_RP_mu} and \eqref{eq:Proof_grad_RP_phi} and then using the standard identity for the variance along with the moments of a standard normal distribution yield the results \eqref{eq:Variance_mu} and \eqref{eq:Variance_phi},
    as required. 
    
    Finally, to prove (iii), we first derive the variances of the Rao-Blackwellized estimator,
\begin{multline}
    \Var_q\left(\Delta^\mathrm{RB}_{\mu_i}(\theta; \lambda)\right) =  3H_i(\mu)^{2\top} \sigma^2 \\ +\frac{3}{4} H_{ii}(\mu)^2\sigma_i^2 + 2G_i(\mu)^2 \label{eq:Variance_mu_RB} 
\end{multline}

\begin{multline}
 \Var_q\left(\Delta^\mathrm{RB}_{\phi_i}(\theta; \lambda)\right)  =  \sigma_i^2\Bigl( 10H_{i}(\mu)^{2\top} \sigma^2 +\\ \frac{37}{2}H_{ii}(\mu)^2 \sigma_i^2 + 10G_i(\mu)^2 \Bigr),\label{eq:Variance_phi_RB}
\end{multline}
where $H_i(\mu)$ is the $i$-th row/column of $H(\mu)$ ($H(\mu)$ is symmetric) and $\sigma^2$ and $H_i(\mu)^2$ are the corresponding vectors squared element-wise. To prove \eqref{eq:Variance_mu_RB} and \eqref{eq:Variance_phi_RB}, we derive the Rao-Blackwellized gradient for the parameters relating to posterior component $i$ using \eqref{eq:ELBOgradRB}. The first step requires finding $h_{-i}(\theta_{(i)})$. To do this, take \eqref{eq:Proof_h_expand} and first remove terms not containing $\theta_i$ (the constant term and terms in the sums not containing $\theta_i$). The Markov blanket for $h(\theta)$ denoted by $\theta_{(i)}$, is defined to be the subset of $\{\theta_j\}_{j=1}^k$ such that $h(\theta)$ is independent of $\theta_j$ for all $j$, conditional on $\theta_{(i)}$. Under the quadratic assumption on the log-joint density, $\theta_{(i)}=\theta$, since the quadratic term in \eqref{eq:Proof_h_expand} has elementwise products between all $\theta_j$ terms. Therefore, $h(\theta)$ cannot be conditionally independent given a subset of $\{\theta_j\}_{j=1}^k$. From this,
    \begin{multline}
        h_{-i}(\theta_{(i)}) = G_i(\mu) (\theta_i-\mu_i) \\ + \sum_{m=1}^k (\theta_i-\mu_i)H_{im}(\mu)(\theta_m-\mu_m).\label{eq:Proof_h_RB}
    \end{multline}
    Combining \eqref{eq:Proof_h_RB} and \eqref{eq:Proof_q_grad}, we can now evaluate \eqref{eq:ELBOgradRB} to be
    \begin{align*}
       \Delta^\mathrm{RB}_{\mu_i}(\theta; \lambda) &=  h_{-i}(\theta_{(i)})\frac{\partial}{\partial\mu_{i}} \log q(\theta_{i}; \mu_{i}, \phi_{i})\\
       &= h_{-i}(\theta_{(i)})\frac{\theta_i - \mu_i}{\sigma_i ^ 2}
    \end{align*}
    and similarly,
    \begin{align*}
        \Delta^\mathrm{RB}_{\phi_i}(\theta; \lambda) &=  h_{-i}(\theta_{(i)})\frac{\partial}{\partial\phi_{i}} \log q(\theta_{i}; \mu_{i}, \phi_{i})\\
       &= h_{-i}(\theta_{(i)}) \left(-1 + \frac{(\theta_i-\mu_i)^2}{\sigma_i^2}\right).
    \end{align*}
    Using the independence between $\theta_i$ and $\theta_j$ for $i\neq j$ from the mean-field assumption, the standard identity for the variance and expressions for the moments of normal random variables yield the marginal variances for the Rao-Blackwellized estimator. Now (iii) follows immediately from (i), (ii), \eqref{eq:Variance_mu_RB} and \eqref{eq:Variance_phi_RB}.
\end{proof}
\end{appendices}

\end{document}